\documentclass{article}
\usepackage{ijcai26}

\usepackage{times}
\usepackage{soul}
\usepackage{url}
\usepackage[hidelinks]{hyperref}
\usepackage[utf8]{inputenc}
\usepackage[small]{caption}
\usepackage{graphicx}
\usepackage{amsmath}
\usepackage{amsthm}
\usepackage{booktabs}
\usepackage{algorithm}
\usepackage{algorithmic}
\usepackage[switch]{lineno}
\usepackage{natbib}
\usepackage{amssymb}
\usepackage{multirow}

\newtheorem{theorem}{Theorem}

\title{Causal-Driven Feature Evaluation for Cross-Domain Image Classification}

\author{
Chen Cheng   
\and
Ang Li\thanks{Corresponding author.}
\\
\affiliations
Florida State University   
\\
\emails
cc24cg@fsu.edu,
al23bp@fsu.edu
}

\begin{document}

\maketitle

\begin{abstract}
Out-of-distribution (OOD) generalization remains a fundamental challenge in real-world classification, where test distributions often differ substantially from training data~\citep{gulrajani2021dg,wang2022domainbed}.
Most existing approaches pursue domain-invariant representations, implicitly assuming that invariance implies reliability.
However, features that are invariant across domains are not necessarily causally effective for prediction.

In this work, we revisit OOD classification from a causal perspective and propose to evaluate learned representations based on their necessity and sufficiency under distribution shift.
We introduce an explicit segment-level framework that directly measures causal effectiveness across domains, providing a more faithful criterion than invariance alone.

Experiments on multi-domain benchmarks demonstrate consistent improvements in OOD performance, particularly under challenging domain shifts, highlighting the value of causal evaluation for robust generalization.
\end{abstract}

\section{Introduction}

Out-of-distribution (OOD) generalization is a central problem in modern machine learning, particularly for classification systems deployed in real-world environments~\citep{gulrajani2021dg,wang2022domainbed}. In practice, models are often trained on data collected under limited conditions, while test-time inputs may come from unseen or shifted distributions~\citep{quionero2009dataset,ben2010theory}
. Such distribution shifts can severely degrade performance, making reliable OOD generalization both practically important and technically challenging.

A widely adopted strategy for OOD classification is to learn domain-invariant representations. The underlying intuition is that features shared across multiple training domains are more likely to generalize to unseen ones. This idea has motivated a large body of work, including invariant risk minimization~\citep{arjovsky2019irm}, distributionally robust optimization~\citep{sagawa2020groupdro}, and domain generalization methods~\citep{gulrajani2021dg,cha2021swad}. While these approaches have demonstrated empirical success, their performance remains inconsistent, especially under severe distribution shifts.

A key limitation of existing methods lies in an implicit assumption: cross-domain invariance implies usefulness for prediction. However, this assumption does not always hold. Some features may remain stable across domains due to shared acquisition processes, background patterns, or preprocessing artifacts, yet contribute little—or even nothing—to the target prediction~\citep{rojas2021causal,scholkopf2021toward}. In other words, domain invariance does not guarantee causal effectiveness. This gap becomes particularly problematic in OOD settings, where spurious but invariant features can dominate the learned representations.

From a causal perspective, robust OOD generalization should rely on features that are not only invariant, but also causally effective for the prediction task across domains~\citep{pearl2009causality,peters2016causal}. That is, changing or removing such features should consistently affect the model’s predictions, regardless of the domain. However, explicitly identifying and evaluating such features remains difficult, especially when models learn high-dimensional and entangled representations~\citep{scholkopf2021toward,locatello2019challenging}.
Motivated by these observations, we aim to answer the following question: how can we directly assess whether a learned feature is causally effective for OOD classification?
\paragraph{Contribution.}
This paper makes the following contributions:
\begin{itemize}
    \item We introduce a causal evaluation framework for OOD classification based on the probability of necessity and sufficiency (PNS)~\citep{pearl1999probabilities}, which directly measures whether learned representation segments are causally effective across domains.
    
    \item We propose a practical two-stage framework that integrates structured representation learning with segment-level causal evaluation under distribution shifts.
    
    \item Experiments on controlled and realistic multi-domain benchmarks demonstrate consistent improvements in OOD generalization, particularly under challenging domain shifts.
\end{itemize}

\section{Related Work}

\paragraph{OOD classification and domain generalization.}
Out-of-distribution (OOD) classification aims to maintain predictive performance when the test distribution differs from training, and is closely related to domain generalization (DG) where multiple training domains are available but the target domain is unseen~\citep{gulrajani2021dg,wang2022domainbed}. A major challenge is that domain shifts can alter spurious correlations, making features that appear stable across training domains still unreliable at test time~\citep{geirhos2020shortcut}
.

\paragraph{Invariant and robust learning.}
A classical line of work seeks invariance across environments to improve OOD generalization.
Invariant Risk Minimization (IRM) encourages predictors whose optimality is invariant across training environments~\citep{arjovsky2019irm}.
Distributionally robust optimization (DRO) methods, such as GroupDRO, optimize the worst-group risk to handle group/domain shifts~\citep{sagawa2020groupdro}.
Beyond objectives, optimization dynamics also matter: SWAD proposes to average weights along a ``flat'' region to improve domain generalization robustness~\citep{cha2021swad}.
More recent works further explore causal and invariant structures for OOD generalization from different perspectives~\citep{mitrovic2021representation,lu2021cdg}.

\paragraph{Causal viewpoints and sufficiency/necessity.}
Causal perspectives motivate evaluating whether a feature is \emph{causally effective} for prediction rather than merely \emph{invariant}~\citep{pearl2009causality,peters2016causal}. In particular, \citep{yang2023pns} introduce invariant learning via the \emph{Probability of Sufficient and Necessary causes} (PNS), arguing that invariance alone may be insufficient: features can be stable across domains yet not causally useful. Their formulation provides a principled way to measure sufficiency/necessity and connect it to OOD risk.

\paragraph{Representation learning for DG.}
Another related direction is to learn representations that separate causal/invariant factors from domain-specific variations~\citep{parascandolo2018learning}.
For example, counterfactual fairness and sequential autoencoding have been explored for domain generalization~\citep{lin2023cdsae}.
IJCAI also includes works on cross-domain feature manipulation/augmentation to bridge domain gaps~\citep{zhang2024csg}.
These approaches support the broader motivation that better factorization of representation can facilitate robust prediction under domain shift.

\section{Methodology}
\label{sec:method}

\subsection{Problem Setup: Multi-Domain OOD Classification}
\label{sec:setup}

Let $\mathcal{X}$ be the input space and $\mathcal{Y}=\{1,\dots,C\}$ be the label space.
We consider $M$ observed source domains $\mathcal{S}=\{S_1,\dots,S_M\}$.
Each domain $S_m$ induces a joint distribution $P_{S_m}(X,Y)$ over $\mathcal{X}\times\mathcal{Y}$.
We denote the domain indicator by a discrete random variable $D\in\{1,\dots,M\}$.

\paragraph{Mixture training distribution.}
During training, samples are drawn from a mixture distribution
\begin{equation}
P_S(X,Y)=\sum_{m=1}^{M}\pi_m P_{S_m}(X,Y),\qquad 
\pi_m>0,\quad \sum_{m=1}^{M}\pi_m=1,
\label{eq:mix_train}
\end{equation}
which can equivalently be written as
$P_S(X,Y)=\mathbb{E}_{D\sim\pi}\!\left[P(X,Y\mid D)\right]$,
where $P(X,Y\mid D=m)=P_{S_m}(X,Y)$.

\paragraph{Unseen target domain.}
At test time, data are drawn from an unseen target domain $T$ with distribution $P_T(X,Y)$.
The target domain is not accessible during training and may differ substantially
from all source domains in appearance, style, or nuisance factors.
Throughout this work, we assume that all domains share the same label space $\mathcal{Y}$,
while the marginal and conditional distributions may vary across domains.

\paragraph{Risk.}
Given a predictor $f:\mathcal{X}\to\Delta^{C-1}$ and a loss function
$\ell:\Delta^{C-1}\times\mathcal{Y}\to\mathbb{R}_+$,
the expected risk on domain $Q$ is defined as
\begin{equation}
\mathcal{R}_Q(f)=\mathbb{E}_{(x,y)\sim P_Q}\!\left[\ell(f(x),y)\right].
\label{eq:risk_def}
\end{equation}
In particular, we also consider the $0$--$1$ risk
\begin{equation}
\mathcal{R}^{0/1}_Q(f)
=\mathbb{E}_{(x,y)\sim P_Q}\!\left[\mathbb{I}\{\arg\max f(x)\neq y\}\right],
\label{eq:risk_01}
\end{equation}
which directly corresponds to classification accuracy.

The goal of out-of-distribution (OOD) classification is to learn a predictor $f$
from samples drawn only from the source domains $\mathcal{S}$,
such that the risk $\mathcal{R}_T(f)$ on unseen target domains is minimized.

\paragraph{Train/test split of domains.}
We partition the observed domains into a set of training domains
$\mathcal{S}_{\mathrm{tr}}$ and a set of held-out test domains
$\mathcal{S}_{\mathrm{te}}$, with $\mathcal{S}_{\mathrm{tr}}\cap\mathcal{S}_{\mathrm{te}}=\emptyset$.
Training data are exclusively drawn from $\mathcal{S}_{\mathrm{tr}}$,
while $\mathcal{S}_{\mathrm{te}}$ is used only for evaluation.

We report per-domain classification accuracy
\begin{equation}
\mathrm{Acc}_{Q}(f)
=\mathbb{E}_{(x,y)\sim P_Q}
\left[\mathbb{I}\{\arg\max f(x)=y\}\right],
\end{equation}
and define the test-domain average performance as
\begin{equation}
\mathrm{Avg}_{\mathrm{te}}(f)
=\frac{1}{|\mathcal{S}_{\mathrm{te}}|}
\sum_{Q\in \mathcal{S}_{\mathrm{te}}}
\mathrm{Acc}_{Q}(f).
\label{eq:test_avg}
\end{equation}

\subsection{Limitations of Observational Invariance for Cross-Domain Causal Reasoning}
\label{sec:why_shared_structure}

A large body of domain generalization and OOD classification methods
aims to learn \emph{domain-invariant representations}.
Let $h:\mathcal{X}\to\mathbb{R}^d$ be a feature extractor and
$g:\mathbb{R}^d\to\Delta^{C-1}$ a classifier, with $f=g\circ h$.
A commonly adopted invariance criterion enforces that
\begin{equation}
P\!\left(h(X)\mid Y, D=i\right)\approx P\!\left(h(X)\mid Y, D=j\right),
\qquad \forall i,j\in\mathcal{S}_{\mathrm{tr}},
\label{eq:invariance}
\end{equation}
which intuitively encourages features predictive of $Y$ to be stable across training domains.

\paragraph{Observational invariance is not interventional invariance.}
The statement in~\eqref{eq:invariance} is purely observational: it constrains
conditional distributions under $P(\cdot\mid D=i)$.
However, causal effectiveness is defined through \emph{interventions} rather than purely observational distributions~\citep{pearl2009causality,bareinboim2016transportability}.
Let $Z=h(X)\in\mathbb{R}^d$ and denote by $Z^{(k)}$ the $k$-th coordinate.
A prototypical feature-level intervention is
\begin{equation}
\mathrm{do}\!\left(Z^{(k)} \leftarrow \tilde{z}\right),
\label{eq:do_coord}
\end{equation}
which conceptually replaces the latent variable $Z^{(k)}$ by an exogenous value $\tilde{z}$ while keeping the remaining coordinates fixed.
Crucially, \eqref{eq:invariance} does \emph{not} constrain the interventional behavior
\begin{equation}
P\!\left(Y \mid \mathrm{do}(Z^{(k)}\!\leftarrow\!\tilde{z}), D=i\right),
\label{eq:interventional_cond}
\end{equation}
nor does it ensure that such interventional responses agree across domains.

\paragraph{Causal evaluation depends on representation coordinates.}
Intervening on $Z^{(k)}$ is only meaningful if $Z^{(k)}$ corresponds to a \emph{well-defined variable}.
Yet the coordinates of a learned representation are not unique, and different parameterizations may encode distinct semantics while yielding identical predictions~\citep{locatello2019challenging,khemakhem2020vae}.

Formally, for any bijective and differentiable mapping
$\phi:\mathbb{R}^d\to\mathbb{R}^d$, one can define an equivalent reparameterization
\begin{equation}
\tilde{h}=\phi\circ h,
\qquad 
\tilde{g}=g\circ \phi^{-1},
\label{eq:repr_map}
\end{equation}
which preserves the input--output behavior
\begin{equation}
\tilde{g}(\tilde{h}(x)) = g(h(x)), \quad \forall x\in\mathcal{X}.
\label{eq:repr_equiv}
\end{equation}
However, an intervention that is ``coordinate-wise'' under $Z$ is generally \emph{not} coordinate-wise under $\tilde{Z}=\phi(Z)$.
Indeed, the intervention $\mathrm{do}(Z^{(k)}\!\leftarrow\!\tilde{z})$ corresponds to a \emph{coupled} transformation in $\tilde{Z}$ unless $\phi$ is coordinate-separable.
Equivalently, the counterfactual prediction map
\begin{equation}
\mathcal{I}_k(Z,\tilde{z}) \triangleq (Z^{(1)},\dots,Z^{(k-1)},\tilde{z},Z^{(k+1)},\dots,Z^{(d)})
\label{eq:intervention_operator}
\end{equation}
is not invariant under general reparameterizations, i.e.,
$\phi(\mathcal{I}_k(Z,\tilde{z}))\neq \mathcal{I}_k(\phi(Z),\cdot)$ in general.
Therefore, causal notions such as sufficiency/necessity/PNS computed by intervening on ``the $k$-th coordinate'' are ill-defined unless the representation admits a fixed coordinate semantics.

\paragraph{Failure of cross-domain causal transfer without aligned structure.}
The ambiguity above becomes critical in the OOD setting.
To formalize ``misaligned semantics'', consider a domain-dependent encoder family
$\{h_i\}_{i=1}^{M}$ (even if implemented by a single network, it may realize different effective parameterizations across domains),
with induced latents $Z_i=h_i(X)\in\mathbb{R}^d$.
Assume that there exists an underlying abstract factor vector $U\in\mathbb{R}^d$ (latent causal/nuisance factors) and domain-dependent bijections $\psi_i$ such that
\begin{equation}
Z_i = \psi_i(U).
\label{eq:domain_latent_map}
\end{equation}
A feature-level causal score computed on domain $i$ for coordinate $k$ implicitly refers to interventions on $Z_i^{(k)}$.
But if $\psi_i$ and $\psi_j$ are not \emph{aligned} (e.g., differ by a non-permutation mixing),
then $Z_i^{(k)}$ and $Z_j^{(k)}$ correspond to different functions of $U$.
Consequently, a coordinate-indexed causal score does not target the same abstract variable across domains.

Concretely, let $\mathrm{CE}_i^{(k)}$ denote any causal-effectiveness functional (e.g., PNS-style) computed using interventions on $Z_i^{(k)}$:
\begin{equation}
\mathrm{CE}_i^{(k)} \triangleq \mathbb{E}_{(x,y)\sim P_{S_i}}
\Big[\Gamma\big(g(Z_i),\, g(\mathcal{I}_k(Z_i,\tilde{z})) ,\, y\big)\Big],
\label{eq:ce_def_general}
\end{equation}
where $\Gamma(\cdot)$ is a score kernel (e.g., indicator of correctness preservation/breaking) and $\tilde{z}$ is sampled from some reference.
Even if $\mathrm{CE}_i^{(k)}$ is well-estimated on $S_i$,
there is no reason that $\mathrm{CE}_i^{(k)}$ should predict the behavior of the ``same $k$'' on $S_j$ unless the coordinate semantics are shared, i.e.,

$\psi_i \approx \psi_j \quad \text{in the sense that} \quad
\mathcal{I}_k(\psi_i(U),\tilde{z}) \approx \psi_i\!\left(\mathcal{I}_k(U,\tilde{u})\right)
\ \text{and similarly for } j$,
\label{eq:alignment_requirement}

for appropriately coupled intervention values $\tilde{z},\tilde{u}$.
Without such alignment, cross-domain causal transfer fails even if observational invariance \eqref{eq:invariance} approximately holds.

\paragraph{Why not intervene in the input space?}
A naive alternative is to define interventions directly on the input $X$,
e.g., at the pixel level.
Let $X\in\mathbb{R}^p$ with $p\gg d$ and consider $\mathrm{do}(X^{(r)}\!\leftarrow\!\tilde{x})$.
Besides being computationally prohibitive (requiring $O(p)$ or worse interventions per instance),
pixel-wise variables lack coherent semantics: there is no guarantee that changing a coordinate $X^{(r)}$ corresponds to an interpretable causal factor.
Formally, if the label depends on a low-dimensional factor $U$ via $Y=\eta(U)$ and $X=\xi(U,D,\epsilon)$,
then intervening on individual pixels does not correspond to interventions on $U$ unless $\xi$ is trivial/invertible at the pixel level,
which is precisely the regime where semantic disentanglement is absent~\citep{chang2018explaining}
.

\paragraph{Implication.}
These considerations suggest that causal evaluation for OOD classification requires representations with semantically aligned latent segments across domains.
Only under such structure do segment-level interventions correspond to the same abstract variables, making causal effectiveness scores transferable to unseen target domains.

\subsection{Shared Latent Coordinate System via Generative Factorization}
\label{sec:latent_coord}

Causal evaluation across domains
is ill-defined without a representation space in which
interventions admit consistent semantic interpretations.
We now describe how such a shared coordinate system
can be induced through a structured generative latent space.

\paragraph{Structured latent representation.}
We learn an encoder $E:\mathcal{X}\to\mathcal{Z}$ that maps inputs to a structured latent space
\begin{equation}
z = E(x) \in \mathcal{Z} \subseteq \mathbb{R}^{K\times d},
\end{equation}
which decomposes into $K$ latent segments (blocks)

\begin{equation}
z = z^{(1)} \oplus z^{(2)} \oplus \cdots \oplus z^{(K)}, 
\qquad z^{(k)} \in \mathbb{R}^{d}.
\label{eq:seg_def}
\end{equation}
Each segment is treated as a \emph{candidate abstract factor}
rather than an arbitrary coordinate subset.
We denote by $z_{\setminus k}$ the concatenation of all segments except $z^{(k)}$.

\paragraph{Generative factorization and semantic anchoring.}
In addition to the encoder, we consider a generator
$G:\mathcal{Z}\to\mathcal{X}$,
forming a reconstruction map $x \mapsto z \mapsto \hat{x}$.
The pair $(E,G)$ induces a \emph{latent coordinate chart}
in which segment-level manipulations in $\mathcal{Z}$
correspond to structured transformations in the input space.

Crucially, the generator serves as a \emph{semantic anchor}:
two latent codes that differ only in segment $k$
produce images that differ in a localized and coherent manner.
This property allows us to interpret operations of the form
\begin{equation}
z \;\longrightarrow\; \tilde{z} = z_{\setminus k} \oplus \bar{z}^{(k)}
\end{equation}
as well-defined interventions on a latent factor,
rather than arbitrary coordinate perturbations.

Generality beyond generative models.
Although we use a generative factorization to induce a shared latent coordinate system, the proposed PNS criterion is not restricted to explicit generative models.
It applies to any representation with semantically aligned segments across domains, e.g., intermediate features of pretrained backbones, where the generator mainly serves as an interpretable realization in controlled settings.

\paragraph{Latent structural causal model.}
We assume an approximate latent data-generating process
\begin{align}
Z &= (Z_{\mathrm{c}}, Z_{\mathrm{s}}), \\
Y &\leftarrow \mathcal{H}(Z_{\mathrm{c}}), \label{eq:latent_label}\\
X &\leftarrow \mathcal{G}(Z_{\mathrm{c}}, Z_{\mathrm{s}}, U, D), \label{eq:latent_image}
\end{align}
where $Z_{\mathrm{c}}$ denotes causally relevant latent factors,
$Z_{\mathrm{s}}$ denotes spurious or domain-dependent factors,
$U$ captures unobserved nuisance variables,
and $D$ is the domain indicator.
The encoder $E$ is not assumed to recover $Z_{\mathrm{c}}$ exactly,
but rather to produce a structured representation
in which causal and spurious information are partially disentangled
across segments.

\paragraph{Semantic alignment across domains.}
The key requirement for causal transfer is not strict invariance,
but \emph{semantic alignment} of latent segments across domains.
Intuitively, the $k$-th segment should correspond to the same
abstract factor (up to reparameterization)
regardless of the domain in which the input is observed.

We formalize this requirement as follows.
An encoder $E$ induces semantically aligned segments
on domains $\mathcal{S}_{\mathrm{tr}}$
if for each segment index $k$
there exists a latent factor $C^{(k)}$
and a family of domain-dependent but \emph{invertible} maps
$\{\psi_D^{(k)}\}_{D\in\mathcal{S}_{\mathrm{tr}}}$
such that
\begin{equation}
z^{(k)} = \psi_D^{(k)}\!\left(C^{(k)}\right),
\qquad \text{for } x\sim P_{S_D},
\label{eq:semantic_align}
\end{equation}
and the conditional relationship
\begin{equation}
P\!\left(Y \mid C^{(k)}, D\right)
\end{equation}
is invariant across domains.

Definition~\ref{def:semantic_alignment} states that,
although the numerical realization of each segment may vary with the domain,
each segment corresponds to a \emph{domain-independent abstract factor}.
This notion is strictly stronger than observational invariance
and directly addresses the representation ambiguity
discussed in Section~\ref{sec:why_shared_structure}.

\paragraph{Why representation structure matters.}
Causal evaluation across domains requires representation components with consistent semantics~\citep{locatello2019challenging}.
This can be induced by a learned generative factorization~\citep{goodfellow2014gan,higgins2017beta} or provided by a pretrained backbone whose intermediate features exhibit stable semantics across domains.

\paragraph{Learning objective.}

Learning objective (when training a representation).
When an explicit representation needs to be learned from scratch, such as in controlled synthetic settings, we induce a shared coordinate system by training an encoder E (and optionally a generator G) with reconstruction and latent-cycle consistency objectives:

\begin{equation}
\min_{E}\ 
\underbrace{\mathbb{E}_{x\sim P_S}\!\left[\|G(E(x)) - x\|_1\right]}_{\text{reconstruction}}
+
\lambda_{\mathrm{lat}}
\underbrace{\mathbb{E}_{z\sim Q}\!\left[\|E(G(z)) - z\|_2^2\right]}_{\text{latent consistency}}.
\label{eq:EG_obj}
\end{equation}
The latent consistency term penalizes domain-specific drift
of latent semantics and promotes reuse of the same latent factors
across domains.

\paragraph{Consequence for causal evaluation.}
Under semantic alignment,
segment-level interventions are well-defined and transferable:
intervening on $z^{(k)}$ corresponds to intervening on the same
abstract factor $C^{(k)}$ across domains.
This property enables meaningful estimation of
cross-domain causal effectiveness scores,
which we introduce in the next section.

\subsection{PNS-style Segment Causal Effectiveness Across Domains}
\label{sec:baseline_cls}

Let $g:\mathcal{Z}\to\Delta^{C-1}$ be a classifier on the full latent $z$.
We write $\hat{y}(z)\triangleq \arg\max_c g_c(z)$.
The baseline classifier is trained by
\begin{equation}
\min_{g}\ \mathbb{E}_{(x,y)\sim P_S}\big[\ell(g(E(x)),y)\big].
\label{eq:baseline_train}
\end{equation}
This baseline provides a stable decision function used later to measure the effect of latent interventions.

\label{sec:pns_segment}

We now quantify whether a segment $z^{(k)}$ is causally effective for predicting $Y$ \emph{across domains}.
We adopt a sufficiency/necessity view inspired by PNS (Pearl) and its invariance-oriented adaptations,
but instantiate the intervention at the \emph{segment level} under a shared latent structure.

\paragraph{Reference distributions for interventions.}
Let $Q_k$ and $Q_{\setminus k}$ be reference distributions for sampling replacement segments.
A practical choice is the aggregated posterior across training domains:
\begin{equation}
Q_k \approx \frac{1}{|\mathcal{S}_{\mathrm{tr}}|}\sum_{D\in\mathcal{S}_{\mathrm{tr}}} P_D(z^{(k)}),
\qquad
Q_{\setminus k} \approx \frac{1}{|\mathcal{S}_{\mathrm{tr}}|}\sum_{D\in\mathcal{S}_{\mathrm{tr}}} P_D(z_{\setminus k}).
\label{eq:Q_def}
\end{equation}
This ensures interventions are not biased toward a single domain.

\paragraph{Intervention operators.}
Define segment-replacement operators:
\begin{align}
\mathcal{I}^{\mathrm{nec}}_{k}(z) &\triangleq z_{\setminus k}\oplus \bar{z}^{(k)},\qquad \bar{z}^{(k)}\sim Q_k,
\label{eq:I_nec}\\
\mathcal{I}^{\mathrm{suf}}_{k}(z) &\triangleq \tilde{z}_{\setminus k}\oplus z^{(k)},\qquad \tilde{z}_{\setminus k}\sim Q_{\setminus k}.
\label{eq:I_suf}
\end{align}
The necessity intervention destroys the original $z^{(k)}$ while keeping the rest; the sufficiency intervention keeps $z^{(k)}$
while randomizing the complement.

\paragraph{Segment sufficiency (SF).}
We measure sufficiency by the probability of \emph{error} when only $z^{(k)}$ is preserved:
\begin{equation}
\mathrm{SF}^{(k)}_D \triangleq
\mathbb{E}_{(x,y)\sim P_D}\ 
\mathbb{E}_{\tilde{z}_{\setminus k}\sim Q_{\setminus k}}
\Big[\mathbb{I}\big[\hat{y}(\mathcal{I}^{\mathrm{suf}}_{k}(E(x)))\neq y\big]\Big].
\label{eq:SF}
\end{equation}
Smaller $\mathrm{SF}^{(k)}_D$ indicates $z^{(k)}$ alone tends to preserve correct predictions.

\paragraph{Segment necessity (NC).}
We measure necessity by the probability of \emph{retained correctness} after destroying $z^{(k)}$:
\begin{equation}
\mathrm{NC}^{(k)}_D \triangleq
\mathbb{E}_{(x,y)\sim P_D}\ 
\mathbb{E}_{\bar{z}^{(k)}\sim Q_k}
\Big[\mathbb{I}\big[\hat{y}(\mathcal{I}^{\mathrm{nec}}_{k}(E(x)))= y\big]\Big].
\label{eq:NC}
\end{equation}
Smaller $\mathrm{NC}^{(k)}_D$ means that overwriting $z^{(k)}$ typically breaks correctness, hence $z^{(k)}$ is necessary.

\paragraph{Explicit segment-level PNS.}
While sufficiency and necessity capture complementary aspects of causal influence,
the classical probability of necessary and sufficient cause (PNS)
requires both conditions to hold simultaneously.
Using the segment-level intervention operators defined above,
we define the explicit PNS for segment $k$ under domain $D$ as
\begin{equation}
\begin{aligned}
\mathrm{PNS}^{(k)}_{D}
\triangleq\;
&\mathbb{E}_{(x,y)\sim P_{D}}\,
 \mathbb{E}_{\substack{\bar z^{(k)}\sim Q_k\\ \tilde z_{\setminus k}\sim Q_{\setminus k}}}
\Big[
\mathbf{1}\Big\{
\begin{array}{l}
\hat y(E(x)) = y,\\
\hat y(\mathcal{I}^{\mathrm{nec}}_{k}(E(x))) \neq y,\\
\hat y(\mathcal{I}^{\mathrm{suf}}_{k}(E(x))) = y
\end{array}
\Big\}
\Big].
\end{aligned}
\label{eq:PNS_explicit}
\end{equation}

\paragraph{Stability / agreement (optional).}
To stabilize estimation, we measure prediction agreement under interventions:
\begin{equation}
\mathrm{M}^{(k)}_D \triangleq
\mathbb{E}_{(x,y)\sim P_D}\ 
\mathbb{E}_{\bar{z}^{(k)}\sim Q_k}
\Big[\mathbb{I}\big[\hat{y}(E(x))=\hat{y}(\mathcal{I}^{\mathrm{nec}}_{k}(E(x)))\big]\Big].
\label{eq:M_def}
\end{equation}

\paragraph{Cross-domain robust aggregation.}
To enforce domain-uniform effectiveness, we use a robust (worst-case) aggregator over training domains:
\begin{equation}
\mathrm{PNS}^{(k)}_{\mathrm{cross}}
\triangleq
-\max_{D\in\mathcal{S}_{\mathrm{tr}}}\mathrm{SF}^{(k)}_D
-\max_{D\in\mathcal{S}_{\mathrm{tr}}}\mathrm{NC}^{(k)}_D
-\lambda_M\max_{D\in\mathcal{S}_{\mathrm{tr}}}\big(1-\mathrm{M}^{(k)}_D\big).
\label{eq:PNS_cross}
\end{equation}
A larger $\mathrm{PNS}^{(k)}_{\mathrm{cross}}$ indicates the segment is sufficient, necessary, and stable
\emph{for every} training domain, which is a stronger requirement than average-case scoring.

\paragraph{Remark (link to causal effectiveness).}
Equations~\eqref{eq:SF}--\eqref{eq:PNS_cross} quantify how the decision changes under controlled segment interventions.
Unlike invariance constraints such as \eqref{eq:invariance}, these quantities are intervention-based and thus closer to causal notions:
they ask whether segment $k$ can \emph{cause} correct predictions when isolated (sufficiency) and whether destroying it \emph{causes}
failure (necessity), uniformly across domains under a shared coordinate system.

\subsection{From PNS Scores to a Two-Stage OOD Classifier}
\label{sec:two_stage}

Our pipeline is explicitly two-stage.

\paragraph{Stage I: learn shared latent structure.}
We learn $E$ and $G$ to produce a domain-aligned latent coordinate system by \eqref{eq:EG_obj}.

\paragraph{Stage II: baseline classifier, PNS scoring, and Top-$K$ retraining.}
We first train baseline $g$ by \eqref{eq:baseline_train}.
Then compute $\mathrm{PNS}^{(k)}_{\mathrm{cross}}$ for each segment $k=1,\dots,K$.
We select the Top-$K^\star$ segments:
\begin{equation}
\mathcal{K}^\star \triangleq \operatorname{TopK}\left(\{\mathrm{PNS}^{(k)}_{\mathrm{cross}}\}_{k=1}^{K},\ K^\star\right),
\qquad
z_{\mathcal{K}^\star}\triangleq \bigoplus_{k\in\mathcal{K}^\star} z^{(k)}.
\label{eq:TopK}
\end{equation}
Finally we retrain a classifier $g^\star$ on the reduced subspace:
\begin{equation}
\min_{g^\star}\ \mathbb{E}_{(x,y)\sim P_S}\big[\ell(g^\star(z_{\mathcal{K}^\star}),y)\big].
\label{eq:TopK_train}
\end{equation}

\paragraph{Interpretation.}
The Top-$K$ selection can be viewed as restricting predictors to causally effective latent segments, thereby controlling hypothesis complexity.

\subsection{Cross-Domain Estimation as Natural Experiments}
\label{sec:cross_domain_est}

Directly estimating causal effects under $P_T$ is infeasible.
Instead we exploit heterogeneity among source domains as natural experiments.

\paragraph{Pairwise shift sensitivity (auxiliary view).}
For domains $(S_i,S_j)$, we can define a shift-based effect proxy
\begin{equation}
\widehat{\mathrm{PNS}}_{i\to j}^{(k)}
=
\mathbb{E}_{(x,y)\sim P_{S_i}}
\Big[
\mathbb{I}\big\{\hat{y}(E(x))\neq \hat{y}(\mathcal{I}^{\mathrm{nec}}_{k}(E(x)))\big\}
\Big],
\label{eq:pair_pns}
\end{equation}
and aggregate across pairs:
\begin{equation}
\widehat{\mathrm{PNS}}^{(k)}=
\frac{1}{M(M-1)}\sum_{i\neq j}\widehat{\mathrm{PNS}}_{i\to j}^{(k)}.
\label{eq:agg_pns}
\end{equation}
Equations~\eqref{eq:pair_pns}--\eqref{eq:agg_pns} highlight that domain shifts provide multiple environments where spurious correlations fluctuate,
helping distinguish causally effective segments.

\paragraph{Connection to our SF/NC.}
Our SF/NC metrics can be viewed as decomposing \eqref{eq:pair_pns} into two complementary intervention regimes
(isolation vs destruction), with robust aggregation \eqref{eq:PNS_cross} to enforce domain-uniformity.

\subsection{Theoretical Analysis of PNS-Guided Selection}
\label{sec:theory}

\subsubsection{Causal vs.\ Spurious Latent Factors}
Assume there exists a latent decomposition
\begin{equation}
C=(C^{\mathrm{causal}},C^{\mathrm{spur}}),
\end{equation}
such that
\begin{align}
Y &= h(C^{\mathrm{causal}}), \label{eq:theory_Y}\\
X &= g(C^{\mathrm{causal}},C^{\mathrm{spur}},D). \label{eq:theory_X}
\end{align}
We allow $C^{\mathrm{spur}}$ to be domain-dependent, which can induce strong but unstable correlations with $Y$ within a domain.

\vspace{0.25em}
\noindent\textbf{Assumptions.}\par
\vspace{0.15em}

\noindent\textbf{Assumption 1 (Causal invariance).}\par
\noindent $P(Y \mid C^{\mathrm{causal}})$ is invariant across domains.\par
\vspace{0.10em}

\noindent\textbf{Assumption 2 (Domain-dependent spuriousness).}\par
\noindent $P(C^{\mathrm{spur}} \mid Y)$ varies across domains.\par
\vspace{0.10em}

\noindent\textbf{Assumption 3 (Encoder faithfulness and block alignment).}\par
\noindent There exists an index set $K_c \subseteq \{1,\ldots,K\}$ such that
blocks $\{z^{(k)}\}_{k\in K_c}$ encode (primarily) $C^{\mathrm{causal}}$ and the
remaining blocks encode (primarily) $C^{\mathrm{spur}}$, and the intervention
distributions $Q_k, Q_{\backslash k}$ in~(24) are non-degenerate and
domain-agnostic.

\subsubsection{Main Result}
\begin{theorem}
\label{thm:pns_separation_v2}
Under Assumptions 1--3, for any block $k$,
the expected cross-domain PNS score satisfies
\begin{equation}
\mathbb{E}\!\left[\mathrm{PNS}^{(k)}_{\mathrm{cross}}\right]
>
\mathbb{E}\!\left[\mathrm{PNS}^{(k')}_{\mathrm{cross}}\right]
\quad
\text{for } k\in \mathcal{K}_{\mathrm{c}},\ k'\notin \mathcal{K}_{\mathrm{c}},
\end{equation}
in the limit of infinite samples and sufficiently expressive classifiers.
\end{theorem}

\paragraph{Proof sketch.}
Under standard causal invariance assumptions, segments encoding causal factors achieve higher PNS scores than spurious ones.
As a result, ranking segments by PNS recovers causally relevant subspaces and improves OOD generalization.
A detailed proof is provided in the Appendix.

\subsection{Algorithm and Computational Cost}
\label{sec:algo}

\begin{algorithm}[t]
\caption{Algorithm 1 Two-stage OOD classification with PNS-guided segment selection (when training representations)
}
\label{alg:pns}
\begin{algorithmic}[1]
\REQUIRE Source domains $\mathcal{S}_{\mathrm{tr}}$, encoder $E$(learned or pretrained), segments $K$, Top-$K^\star$.
\STATE If learning representations from scratch, train $E$ via \eqref{eq:EG_obj}.
\STATE Train baseline classifier $g$ via \eqref{eq:baseline_train}.
\FOR{$k=1$ to $K$}
  \FOR{each domain $D\in\mathcal{S}_{\mathrm{tr}}$}
    \STATE Estimate $\mathrm{SF}^{(k)}_D$ by sampling $\tilde{z}_{\setminus k}\sim Q_{\setminus k}$ and evaluating \eqref{eq:SF}.
    \STATE Estimate $\mathrm{NC}^{(k)}_D$ by sampling $\bar{z}^{(k)}\sim Q_{k}$ and evaluating \eqref{eq:NC}.
  \ENDFOR
  \STATE Compute $\mathrm{PNS}^{(k)}_{\mathrm{cross}}$ by \eqref{eq:PNS_cross}.
\ENDFOR
\STATE Select $\mathcal{K}^\star=\operatorname{TopK}(\mathrm{PNS}^{(k)}_{\mathrm{cross}},K^\star)$.
\STATE Retrain $g^\star$ on $z_{\mathcal{K}^\star}$ via \eqref{eq:TopK_train}.
\RETURN $f^\star(x)=g^\star(E(x)_{\mathcal{K}^\star})$.
\end{algorithmic}
\end{algorithm}

\paragraph{Cost.}
Let $N$ be the number of samples used for PNS estimation per domain and $R$ the number of Monte-Carlo draws for interventions.
Computing SF/NC across $K$ segments costs $O(|\mathcal{S}_{\mathrm{tr}}|\cdot K\cdot N\cdot R)$ forward passes through $g$,
which is feasible since $K$ is small (segment-level rather than per-dimension).


\section{Experiments}
\label{sec:experiments}

This section evaluates the proposed explicit segment-level PNS metric.
We primarily conduct experiments on a carefully controlled multi-domain MNIST benchmark, which enables precise characterization of necessity and sufficiency under predefined domain shifts.
In addition, we report results on the PACS benchmark to examine whether the proposed segment-level PNS criterion transfers to a standard and more realistic domain generalization setting.

\begin{table*}[t]
\centering
\small
\begin{tabular}{l|cccc|cccc|cc}
\toprule
Method &
\multicolumn{4}{c|}{\textbf{Train Domains}} &
\multicolumn{4}{c|}{\textbf{Test Domains}} 
 \\
& clean & noise & blur & contrast
& rotate & invert & thick & thin &  Avg & Min\\
\midrule
GroupDRO    & 99.10 & 98.79 & 98.00 & 98.93 & 96.36 & 29.76 & 95.97 & 60.00 & 70.52 & 29.76\\
MMD    & 98.38 & 99.96 & 99.57 & 96.80 & 96.52 & 34.21 & 93.73 & 52.43  & 69.22 & 34.21\\
CDANN   & 96.23 & 97.84 & 96.57 & 99.78 & 93.59 & 26.78 & 94.41 & 66.29 & 70.26 & 26.78\\
IRM         & 99.47 & 99.19 & 99.39 & 99.47 & \textbf{97.24} & 38.25 & \textbf{97.38} & 55.71 & 72.15 & 38.25 \\
CaSN(IRM) & 95.58 & 99.02 & 98.33 & 98.78 & 97.13 & 39.25 & 96.43 & 58.99 & 72.95 & 39.25 \\
MLDG      & 98.90 & 97.39 & 94.05 & 94.76 & 96.04 & 41.56 & 97.37 & 27.39 & 65.09 & 27.39 \\
baseline (ERM) & 98.40 & 97.64 & 98.05 & 97.77 & 92.75 & 9.79 & 90.67 & 38.41 & 57.91 & 9.79 \\
\midrule
Ours (Top-2) & 96.66 & 93.38 & 96.83 & 96.60 & 95.86 & 83.42 & 94.48 & 75.29 & 87.26 & 75.29\\

\textbf{Ours (Top-3)} &
96.78 & 94.03 & 96.86 & 96.41 &
95.52 & \textbf{85.43} & 94.73 & \textbf{75.83} &
\textbf{87.38} & \textbf{75.83}\\
Ours (Top-4) & 96.77 & 94.17 & 96.85 & 96.50 & 95.64 & 80.93 & 94.84 & 75.12 & 86.88 & 75.12\\
\bottomrule
\end{tabular}
\caption{Domain generalization accuracy (\%). Train domains = clean, noise, blur, contrast. Test domains = rotate, invert, thick, thin. 
Avg is the average accuracy over the 4 unseen domains. Min is the minimize accuracy over the 4 unseen domains.}
\label{tab:dg_results}
\end{table*}

\begin{table*}[t]
\centering
\small
\setlength{\tabcolsep}{5pt}
\renewcommand{\arraystretch}{1.1}
\begin{tabular}{l c c c c c c c c}
\toprule
\textbf{Train $\rightarrow$ Test}
& ERM
& IRM
& GroupDRO
& MLDG
& MMD
& CDANN
& CaSN(IRM)
& PNS (Ours) \\
\midrule
A+C $\rightarrow$ P+S
& 0.8678 & 0.8987 & 0.8993 & 0.9114 & 0.9258 & 0.9035 & 0.9125 & \textbf{0.9138} \\
A+P $\rightarrow$ C+S
& 0.7233 & 0.6486 & 0.6457 & 0.7118 & 0.7424 & 0.6550 & 0.7556 & \textbf{0.8687} \\
A+S $\rightarrow$ C+P
& 0.9071 & 0.8948 & 0.8958 & 0.9109 & 0.8973 & 0.8715 & 0.9142 & \textbf{0.9639} \\
C+P $\rightarrow$ A+S
& 0.8345 & 0.8271 & 0.8238 & 0.8298 & 0.8694 & 0.8446 & 0.8619 & \textbf{0.8989} \\
C+S $\rightarrow$ A+P
& 0.8479 & 0.7632 & 0.7557 & 0.8345 & 0.7421 & 0.7351 & 0.8616 & \textbf{0.8893} \\
P+S $\rightarrow$ A+C
& 0.8300 & 0.8032 & 0.8012 & 0.8425 & 0.8548 & 0.7988 & 0.8370 & \textbf{0.9487} \\
\midrule
\textbf{Mean}
& 0.8351 & 0.8059 & 0.8036 & 0.8402 & 0.8386 & 0.8014 & 0.8571 & \textbf{0.9139} \\
\bottomrule
\end{tabular}
\caption{Results on 2→2 domain generalization over the PACS benchmark.
In the 2→2 setting, models are trained on two source domains and evaluated on the remaining two target domains, which are unseen during training.
A, C, P, and S denote Art, Cartoon, Photo, and Sketch, respectively.
Average accuracy (Avg) on PACS 2$\rightarrow$2 domain generalization. 
Each row corresponds to a train–test split and each column to a method.
PNS consistently achieves the best performance across all splits.}
\label{tab:pacs_2to2_avg_main}
\end{table*}

\subsection{Dataset Construction}
\label{sec:dataset}

We construct a controlled multi-domain image dataset based on the MNIST handwritten digit benchmark~\citep{lecun1998mnist}.
Starting from the original MNIST train/test split, we generate multiple domains by applying predefined image-level transformations that preserve digit semantics while inducing diverse appearance shifts.

\paragraph{Multi-domain MNIST.}
Let $(x,y)$ denote an MNIST image--label pair.
For each domain $d \in \mathcal{D}$, we define a deterministic transformation $T_d(\cdot)$ and construct
\[
x_d = T_d(x), \qquad y_d = y,
\]
where the label remains unchanged across domains.
The domain set $\mathcal{D}$ includes clean, noise, blur, contrast, rotate, invert, thick, and thin, covering a range of mild to severe distributional shifts. Implementation details of the transformations are provided in the Appendix.

\paragraph{Additional benchmark (PACS).}
In addition to the controlled MNIST setting, we evaluate our method on the PACS benchmark~\citep{li2017deeper}, which consists of four visual domains: Art, Cartoon, Photo, and Sketch.
We follow the standard 2$\rightarrow$2 domain generalization protocol and adopt the commonly used train--test splits from prior work, without additional preprocessing.

\subsection{Experimental Setup}

We evaluate the proposed method on both multi-domain MNIST and PACS under distributional shifts.
In all experiments, inputs are encoded into latent representations, which are decomposed into fixed segments for causal analysis.
During PNS estimation, model parameters are kept fixed to isolate the effect of segment-level interventions.
Necessity and sufficiency interventions follow the definitions in Section~\ref{sec:method} and are applied consistently across all settings.

\subsection{Evaluation Metric}
\label{sec:metric}

For each segment $k$, we estimate the explicit PNS defined as:
\begin{equation}
\begin{aligned}
\mathrm{PNS}^{(k)}_{D}
\triangleq\;
&\mathbb{E}_{(x,y)\sim P_{D}}\,
 \mathbb{E}_{\substack{\bar z^{(k)}\sim Q_k\\ \tilde z_{\setminus k}\sim Q_{\setminus k}}}
\Big[
\mathbf{1}\Big\{
\begin{array}{l}
\hat y(E(x)) = y,\\
\hat y(\mathcal{I}^{\mathrm{nec}}_{k}(E(x))) \neq y,\\
\hat y(\mathcal{I}^{\mathrm{suf}}_{k}(E(x))) = y
\end{array}
\Big\}
\Big].
\end{aligned}
\end{equation}

This metric assigns a high value to segments that are simultaneously necessary for maintaining the original prediction and sufficient to recover it when isolated.

\paragraph{Task-level metric.}
We report standard classification accuracy on target domains.
For MNIST we report per-domain accuracy as well as Avg/Min over unseen domains; for PACS 2$\rightarrow$2 we report per-split accuracy and the Mean over all splits.

\subsection{Results and Analysis}
\label{sec:analysis}

Table~\ref{tab:dg_results} and Table~\ref{tab:pacs_2to2_avg_main} summarize the results on the multi-domain MNIST and PACS benchmarks, respectively.
These two evaluations provide complementary evidence for the effectiveness of the proposed explicit PNS metric: MNIST offers a controlled setting for causal analysis under predefined shifts, while PACS evaluates robustness under realistic cross-domain variation.

\paragraph{Overall performance.}
Across both benchmarks, our method consistently achieves the strongest OOD generalization performance.
On multi-domain MNIST, \textbf{Ours (Top-3)} attains the highest average accuracy of \textbf{87.38\%} on unseen target domains, substantially outperforming representative baselines such as IRM, GroupDRO, MMD, and CDANN.
On PACS 2$\rightarrow$2 domain generalization, PNS achieves the best mean accuracy of \textbf{0.9139} across all train--test splits.
These results indicate that selecting representation segments based on explicit joint necessity and sufficiency yields robust improvements under both controlled and realistic domain shifts.

\paragraph{Performance under challenging shifts.}
The advantage of the proposed method is most pronounced under severe distributional shifts.
On multi-domain MNIST, baseline methods degrade significantly on challenging domains such as \emph{invert} and \emph{thin}, while PNS maintains strong performance.
A similar trend is observed on PACS for difficult transfers such as A+P$\rightarrow$C+S and P+S$\rightarrow$A+C.
This suggests that explicit PNS effectively filters out invariant but non-causal features that fail under strong shifts.

\paragraph{Robustness and segment selection.}
Selecting a small number of top-ranked segments already yields strong performance.
Across both benchmarks, Top-3 achieves the best balance between robustness and generalization, while selecting fewer segments leads to mild underfitting and including additional segments introduces marginal degradation.
This stability indicates that PNS ranks segments by causal relevance and that only a small subset is sufficient to capture invariant predictive structure shared across domains.

\section{Conclusion and Future Work}
\label{sec:conclusion}

We proposed an explicit segment-level formulation of the PNS for causal attribution in learned representations.
By enabling targeted intervention-based evaluation, the proposed criterion identifies causally effective segments that support robust OOD generalization.

Experiments on both controlled multi-domain MNIST and the PACS benchmark demonstrate consistent robustness improvements under challenging domain shifts, indicating that explicit causal evaluation provides a stronger criterion than invariance alone.
Overall, this work bridges causal inference and representation learning by offering a practical and interpretable framework for segment-level causal analysis.

\bibliographystyle{named}
\bibliography{ijcai26}

\clearpage
\appendix
\title{Supplementary Material for\\
\emph{Causal-Driven Feature Evaluation for Cross-Domain Image Classification}}

\author{}
\date{}

\maketitle

\section*{Appendix}

This document provides supplementary materials that complement the main paper,
including detailed dataset construction and extended theoretical proofs.

\section{Dataset Construction Details}

\subsection{MNIST Multi-Domain Benchmark}

We construct a controlled multi-domain benchmark based on the MNIST handwritten digit dataset~\citep{lecun1998mnist}.
All data are generated locally from the original MNIST IDX files without external preprocessing or online augmentation.

\paragraph{Source data.}
Each sample consists of an image--label pair $(x,y)$, where
$x \in \mathbb{R}^{28 \times 28}$ is a grayscale digit image and
$y \in \{0,\dots,9\}$ denotes the digit class.
Images are decoded in grayscale format and resized to $32\times32$ using bilinear interpolation.
All images are stored as single-channel grayscale PNG files.

\paragraph{Domain transformations.}
Starting from a fixed MNIST split (train or test), we generate eight domains using deterministic image-level transformations.
For each domain $d$, we define a transformation $T_d(\cdot)$ and construct $x_d = T_d(x)$ while keeping the label unchanged.

The set of domains includes:
\begin{itemize}
    \item \textbf{clean}: identity transformation.
    \item \textbf{noise}: additive Gaussian noise with standard deviation $0.25$, followed by clipping to $[0,1]$.
    \item \textbf{blur}: Gaussian blur with radius $1.2$.
    \item \textbf{contrast}: contrast reduction with factor $0.4$.
    \item \textbf{rotate}: rotation by $20^\circ$ with bilinear interpolation and zero padding.
    \item \textbf{invert}: pixel-wise intensity inversion.
    \item \textbf{thick}: morphological thickening using a max filter with kernel size $3$.
    \item \textbf{thin}: light stroke thinning implemented via Gaussian blur, min filtering, and contrast reduction.
\end{itemize}

All transformations preserve the semantic content and class labels of the digits while inducing distinct appearance shifts.

\paragraph{Data organization.}
For each domain, images are stored in a separate directory, and labels are recorded in a CSV file mapping image filenames to digit classes.
The same domain transformations are applied independently to the training and test splits.

\subsection{PACS Benchmark}

In addition to MNIST, we evaluate our method on the PACS benchmark~\citep{li2017deeper},
which consists of four visual domains: Art, Cartoon, Photo, and Sketch.
We follow the standard $2\rightarrow2$ domain generalization protocol,
where models are trained on two source domains and evaluated on the remaining two unseen target domains.
All train--test splits and preprocessing follow prior work, and no additional domain-specific modifications are introduced.

\section{Extended Theoretical Proofs}

\begin{figure}
    \centering
    \includegraphics[width=0.5\linewidth]{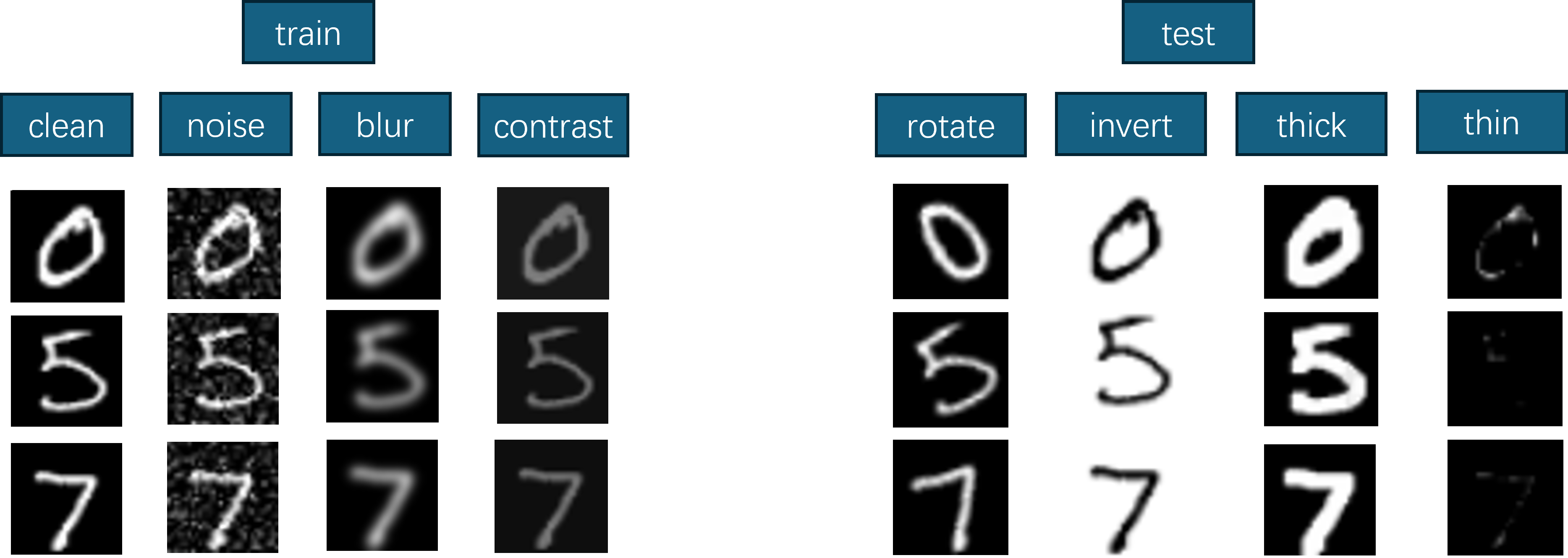}
    \caption{Examples of training and test domain transformatins used in our benchmark.}
    \label{fig:placeholder}
\end{figure}

\subsection{Proof Sketch (Expanded)}

We provide an expanded proof sketch for the separation between causal and non-causal representation segments
under the proposed probability of necessity and sufficiency (PNS) criterion.

\paragraph{(i) Sufficiency separation.}
For a causal segment $k \in \mathcal{K}_{\mathrm{c}}$, the segment representation $z^{(k)}$
encodes information that directly determines the target label $Y$.
As a result, predictions based on $\hat{y}(\tilde{z}_{\setminus k} \oplus z^{(k)})$
remain correct with high probability even when all other segments are randomized.
This implies that the sufficiency failure score remains small for causal segments across all training domains.
In contrast, for a non-causal segment $k' \notin \mathcal{K}_{\mathrm{c}}$,
the segment alone is insufficient to recover $Y$, leading to consistently higher sufficiency failure.

\paragraph{(ii) Necessity separation.}
For a causal segment $k \in \mathcal{K}_{\mathrm{c}}$, overwriting $z^{(k)}$
destroys part of the causal information required for prediction,
resulting in a substantial drop in prediction accuracy.
Consequently, the retained correctness under necessity intervention is low.
For non-causal segments, overwriting the segment does not consistently affect prediction
as long as the causal segments remain intact, leading to comparatively higher retained correctness.

\paragraph{(iii) Worst-case aggregation across domains.}
Spurious segments may appear predictive in individual domains but fail under others due to domain-specific correlations.
By aggregating necessity and sufficiency scores using a worst-case criterion across training domains,
any segment that fails to satisfy either condition in at least one domain is penalized.
Causal segments, which remain effective across domains by assumption, are preserved under this aggregation.

\paragraph{(iv) Consequence for Top-$K$ selection.}
Since causal segments achieve higher PNS scores in expectation,
ranking segments by the proposed cross-domain PNS criterion asymptotically recovers the causal subset.
Restricting predictors to the selected top-$K$ segments therefore improves out-of-distribution generalization.

\section{Additional PACS Results}
\label{sec:appendix_pacs}

\subsection{Detailed Results for PACS $2\rightarrow2$ Domain Generalization}

Table~\ref{tab:pacs_2to2_irm_pns} reports detailed per-split results on the PACS benchmark
under the standard $2\rightarrow2$ domain generalization protocol.
For each train--test configuration, we report accuracy on the two unseen target domains
(\textbf{Acc$_1$} and \textbf{Acc$_2$}) as well as their average.

\begin{table}[t]
\centering
\small
\setlength{\tabcolsep}{5pt}
\renewcommand{\arraystretch}{1.1}
\begin{tabular}{l l c c c}
\toprule
\textbf{Method} & \textbf{Train $\rightarrow$ Test} & \textbf{Acc$_1$} & \textbf{Acc$_2$} & \textbf{Avg} \\
\midrule
\multirow{6}{*}{ERM (Base)}
& A+C $\rightarrow$ P+S & 0.9934 & 0.7422 & 0.8678 \\
& A+P $\rightarrow$ C+S & 0.7487 & 0.6979 & 0.7233 \\
& A+S $\rightarrow$ C+P & 0.8298 & 0.9844 & 0.9071 \\
& C+P $\rightarrow$ A+S & 0.9248 & 0.7442 & 0.8345 \\
& C+S $\rightarrow$ A+P & 0.8892 & 0.8066 & 0.8479 \\
& P+S $\rightarrow$ A+C & 0.9189 & 0.7410 & 0.8300 \\
\midrule

\multirow{6}{*}{IRM}
& A+C $\rightarrow$ P+S & 0.9934 & 0.8040 & 0.8987 \\
& A+P $\rightarrow$ C+S & 0.6476 & 0.6495 & 0.6486 \\
& A+S $\rightarrow$ C+P & 0.8016 & 0.9880 & 0.8948 \\
& C+P $\rightarrow$ A+S & 0.9219 & 0.7322 & 0.8271 \\
& C+S $\rightarrow$ A+P & 0.8677 & 0.6587 & 0.7632 \\
& P+S $\rightarrow$ A+C & 0.8828 & 0.7235 & 0.8032 \\
\midrule

\multirow{6}{*}{GroupDRO}
& A+C $\rightarrow$ P+S & 0.9940 & 0.8045 & 0.8993 \\
& A+P $\rightarrow$ C+S & 0.6433 & 0.6480 & 0.6457 \\
& A+S $\rightarrow$ C+P & 0.8089 & 0.9826 & 0.8958 \\
& C+P $\rightarrow$ A+S & 0.9204 & 0.7272 & 0.8238 \\
& C+S $\rightarrow$ A+P & 0.8574 & 0.6539 & 0.7557 \\
& P+S $\rightarrow$ A+C & 0.8892 & 0.7133 & 0.8012 \\
\midrule

\multirow{6}{*}{MLDG}
& A+C $\rightarrow$ P+S & 0.9876 & 0.8352 & 0.9114 \\
& A+P $\rightarrow$ C+S & 0.7461 & 0.6774 & 0.7118 \\
& A+S $\rightarrow$ C+P & 0.8325 & 0.9893 & 0.9109 \\
& C+P $\rightarrow$ A+S & 0.9439 & 0.7156 & 0.8298 \\
& C+S $\rightarrow$ A+P & 0.8937 & 0.7752 & 0.8345 \\
& P+S $\rightarrow$ A+C & 0.8828 & 0.8022 & 0.8425 \\
\midrule

\multirow{6}{*}{MMD}
& A+C $\rightarrow$ P+S & 0.9892 & 0.8624 & 0.9258 \\
& A+P $\rightarrow$ C+S & 0.7785 & 0.7063 & 0.7424 \\
& A+S $\rightarrow$ C+P & 0.8365 & 0.9582 & 0.8973 \\
& C+P $\rightarrow$ A+S & 0.9672 & 0.7715 & 0.8694 \\
& C+S $\rightarrow$ A+P & 0.8390 & 0.6451 & 0.7421 \\
& P+S $\rightarrow$ A+C & 0.8659 & 0.8436 & 0.8548 \\
\midrule

\multirow{6}{*}{CDANN}
& A+C $\rightarrow$ P+S & 0.9676 & 0.8394 & 0.9035 \\
& A+P $\rightarrow$ C+S & 0.6429 & 0.6670 & 0.6550 \\
& A+S $\rightarrow$ C+P & 0.7947 & 0.9483 & 0.8715 \\
& C+P $\rightarrow$ A+S & 0.9264 & 0.7627 & 0.8446 \\
& C+S $\rightarrow$ A+P & 0.8316 & 0.6385 & 0.7351 \\
& P+S $\rightarrow$ A+C & 0.8657 & 0.7319 & 0.7988 \\
\midrule

\multirow{6}{*}{CaSN (IRM)}
& A+C $\rightarrow$ P+S & 0.9960 & 0.8290 & 0.9125 \\
& A+P $\rightarrow$ C+S & 0.7838 & 0.7274 & 0.7556 \\
& A+S $\rightarrow$ C+P & 0.8485 & 0.9799 & 0.9142 \\
& C+P $\rightarrow$ A+S & 0.9446 & 0.7792 & 0.8619 \\
& C+S $\rightarrow$ A+P & 0.9098 & 0.8134 & 0.8616 \\
& P+S $\rightarrow$ A+C & 0.9337 & 0.7402 & 0.8370 \\
\midrule

\multirow{6}{*}{\textbf{PNS (Ours)}}
& A+C $\rightarrow$ P+S & \textbf{0.9940} & \textbf{0.8335} & \textbf{0.9138} \\
& A+P $\rightarrow$ C+S & \textbf{0.8929} & \textbf{0.8445} & \textbf{0.8687} \\
& A+S $\rightarrow$ C+P & \textbf{0.9356} & \textbf{0.9922} & \textbf{0.9639} \\
& C+P $\rightarrow$ A+S & \textbf{0.9766} & \textbf{0.8213} & \textbf{0.8989} \\
& C+S $\rightarrow$ A+P & \textbf{0.9487} & \textbf{0.8299} & \textbf{0.8893} \\
& P+S $\rightarrow$ A+C & \textbf{0.9712} & \textbf{0.9262} & \textbf{0.9487} \\
\bottomrule
\end{tabular}
\caption{Detailed $2\rightarrow2$ domain generalization results on the PACS benchmark.
A, C, P, and S denote Art, Cartoon, Photo, and Sketch, respectively.
Acc$_1$ and Acc$_2$ correspond to accuracy on the two unseen target domains,
and Avg denotes their mean.}
\label{tab:pacs_2to2_irm_pns}
\end{table}

\end{document}